\documentclass[sigconf]{acmart}

\usepackage{multirow}
\usepackage{colortbl}
\usepackage{amsmath}
\usepackage{bm}

\AtBeginDocument{%
  }

\acmDOI{}
\acmISBN{}
\copyrightyear{2027}
\acmYear{2027}
\acmConference[ASP-DAC '27]{32nd Asia and South Pacific Design Automation Conference}{January 25--28, 2027}{Tokyo, Japan}

\begin{document}

\title[ABNAT: Attention-Based Noise-Aware Training for Robust Transformers on Analog CIM Systems]{ABNAT: Attention-Based Noise-Aware Training for Robust Transformers on Analog Compute-in-Memory Systems}


\author{Wenshuai Yao}
\authornote{Wenshuai Yao and Wenyong Zhou contributed equally to this work.}
\affiliation{%
  \department{School of Integrated Circuits}
  \institution{Peking University}
  \city{Beijing}
  \country{China}}

\author{Wenyong Zhou}
\authornotemark[1]
\affiliation{%
  \department{Department of Electrical and Computer Engineering}
  \institution{The University of Hong Kong}
  \city{Hong Kong}
  \country{China}}

\author{Hanyong Shao}
\affiliation{%
  \department{School of Integrated Circuits}
  \institution{Peking University}
  \city{Beijing}
  \country{China}}

\author{Zhiyuan Ning}
\affiliation{%
  \department{School of Integrated Circuits}
  \institution{Peking University}
  \city{Beijing}
  \country{China}}

\author{Yizhe Chen}
\affiliation{%
  \department{School of Integrated Circuit Science and Engineering}
  \institution{Beihang University}
  \city{Beijing}
  \country{China}}

\author{Yuannuo Feng}
\affiliation{%
  \department{School of Integrated Circuit Science and Engineering}
  \institution{Beihang University}
  \city{Beijing}
  \country{China}}

\author{Kechao Tang}
\authornote{Corresponding author: Kechao Tang.}
\email{tkch@pku.edu.cn}
\affiliation{%
  \department{School of Integrated Circuits}
  \institution{Peking University}
  \city{Beijing}
  \country{China}}
\affiliation{%
  \institution{Beijing Advanced Innovation Center for Integrated Circuits}
  \city{Beijing}
  \country{China}}

\author{Ru Huang}
\affiliation{%
  \department{School of Integrated Circuits}
  \institution{Peking University}
  \city{Beijing}
  \country{China}}
\affiliation{%
  \institution{Beijing Advanced Innovation Center for Integrated Circuits}
  \city{Beijing}
  \country{China}}

\renewcommand{\shortauthors}{Yao et al.}

\makeatletter
\def\@mkauthors@iii{%
  \gdef\@currentauthors{}%
  \gdef\@currentaffiliation{}%
  \global\setbox\mktitle@bx=\vbox{%
    \noindent\unvbox\mktitle@bx\par\medskip
    \begingroup
      \centering\normalfont
      \parindent=0pt\relax
      \parskip=0pt\relax
      {\fontsize{11.5}{13.5}\selectfont
        \mbox{Wenshuai Yao\textsuperscript{1\dag},
          Wenyong Zhou\textsuperscript{2\dag},
          Hanyong Shao\textsuperscript{1},
          Zhiyuan Ning\textsuperscript{1},
          Yizhe Chen\textsuperscript{3},
          Yuannuo Feng\textsuperscript{3},}\par
        \mbox{Kechao Tang\textsuperscript{1,4*},
          Ru Huang\textsuperscript{1,4}}\par}
      \vspace{2pt}
      {\fontsize{10}{12}\selectfont
        \mbox{\textsuperscript{1}School of Integrated Circuits, Peking University, Beijing, China}\par
        \mbox{\textsuperscript{2}Department of Electrical and Computer Engineering, The University of Hong Kong, Hong Kong}\par
        \mbox{\textsuperscript{3}School of Integrated Circuit Science and Engineering, Beihang University, Beijing, China}\par
        \mbox{\textsuperscript{4}Beijing Advanced Innovation Center for Integrated Circuits, Beijing, China}\par}
      \vspace{1.5pt}
      {\fontsize{9.5}{11.5}\selectfont
        \textsuperscript{\dag} Equal contribution.\quad
        \textsuperscript{*} Corresponding author:
        \href{mailto:tkch@pku.edu.cn}{\nolinkurl{tkch@pku.edu.cn}}.\par}
    \endgroup
    \par\bigskip}}
\let\@authornotes\@empty
\makeatother

\begin{abstract}
Transformer-based models have become the backbone of a wide range of tasks, but their intensive computation and data movement incur substantial energy overhead. Analog compute-in-memory (CIM) offers an energy-efficient alternative, yet its inherent hardware non-idealities perturb stored weights and can significantly degrade model performance. Although existing noise-aware training (NAT) methods improve robustness by injecting noise during training, they largely overlook the critical role of attention mechanisms in Transformer inference.
To bridge this gap, we analyze how CIM-induced noise reshapes attention in Transformers and identify two key observations: (1) the divergence between clean and noisy attention maps increases substantially in deeper layers, and (2) conventional NAT can remain overconfident even when its attention routes deviate markedly from the clean reference. Motivated by these findings, we propose ABNAT, an attention-based noise-aware training framework that teaches the noisy student to preserve the clean teacher's headwise attention routes rather than merely reproduce answers, without hardware changes or inference overhead; selective variants further reduce training cost.
Extensive experiments across vision, language, and multimodal Transformers show that ABNAT consistently improves robustness over corresponding NAT baselines, improving noise-degraded accuracy by up to 11.28\% and reducing attention deviation by up to 85.77\%.
\end{abstract}




\maketitle

\setlength{\abovedisplayskip}{2pt plus 0.5pt minus 0.5pt}
\setlength{\belowdisplayskip}{2pt plus 0.5pt minus 0.5pt}
\setlength{\abovedisplayshortskip}{0pt plus 0.5pt}
\setlength{\belowdisplayshortskip}{1.5pt plus 0.5pt minus 0.5pt}

\section{Introduction}
\label{sec:introduction}

Transformers underpin modern vision, language, and multimodal models, but their intensive matrix operations impose substantial computation, memory, and data-movement costs. Analog Compute-in-Memory (CIM) performs matrix-vector multiplications (MVMs) in situ and therefore offers an energy-efficient deployment substrate~\cite{sebastian2020memory, ielmini2018memory, yao2020fully}. However, device variation, write noise, and other analog non-idealities perturb stored weights and can severely degrade Transformer inference.
\begin{figure}[!t]
\centering
\includegraphics[width=1.0\columnwidth]{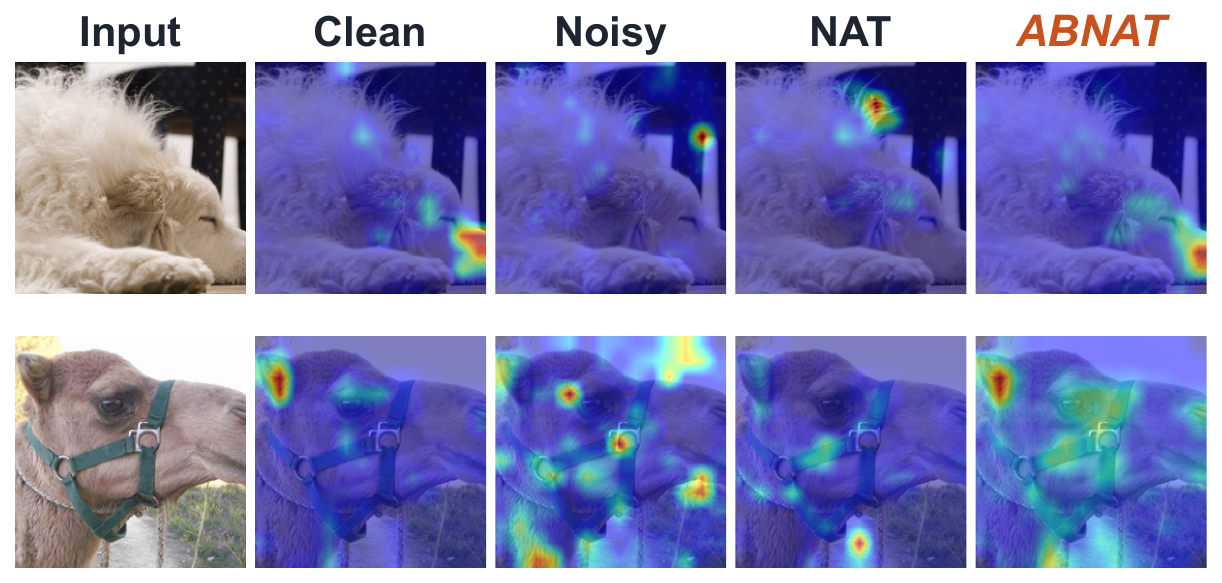}
\caption{Attention maps from clean inference and from the noisy model, NAT, and ABNAT under CIM noise.}
\label{fig:attn_vis}
\end{figure}

To mitigate such degradation, noise-aware training (NAT) methods incorporate hardware noise injection during training to improve model robustness~\cite{joshi2020accurate, zhou2020noisymachines, rasch2023hardwareaware}. While effective for conventional convolutional neural networks (CNNs), these approaches treat the model as a black box and optimize solely for correct output predictions, overlooking the distinctive role of attention mechanisms in Transformer inference. This output-centric paradigm leaves a critical gap: the internal semantic reasoning pathways of Transformers remain fragile under CIM noise, even when final predictions appear correct.

For clear spatial visualization of attention routes, we conduct a detailed analysis of Vision Transformers (ViTs)~\cite{dosovitskiy2021vit}. \textbf{(L)} Output-centric NAT misses semantic attention routes. Existing NAT methods supervise only final predictions, allowing models to produce correct answers via incorrect attention routes. Figure~\ref{fig:attn_vis} shows severe attention distortion under NAT, motivating two structural observations. \textbf{(O1)} CIM noise accumulates attention drift across depth. Weight perturbations disrupt attention concentration at each layer, and errors accumulate with depth. Figure~\ref{fig:layerwise_drift} shows that this drift grows with noise severity and becomes concentrated in late blocks, motivating targeted route preservation. \textbf{(O2)} Head sensitivity is nonuniform within blocks. Figure~\ref{fig:sensitivity_heatmap} further reveals marked head-to-head sensitivity variation, so uniform alignment over all layers is unnecessarily costly without commensurate accuracy gains.

Motivated by these observations, we propose \textbf{ABNAT} (\textbf{A}ttention-\textbf{B}ased \textbf{N}oise-\textbf{A}ware \textbf{T}raining), an attention-based knowledge distillation (KD) framework for preserving semantic routes under CIM noise. ABNAT uses a clean, frozen copy of the target deployment model as the teacher and its noise-perturbed counterpart as the student. Rather than relying only on answer-level supervision, ABNAT transfers the content understanding encoded in the teacher's attention routes by aligning teacher and student attention distributions. Figure~\ref{fig:attn_vis} shows that ABNAT restores noise-distorted attention maps toward the clean reference across diverse inputs, indicating more faithful preservation of semantic attention routes. We further develop head- and block-selective variants that retain near-full robustness at lower training cost, and validate ABNAT across vision, language, and vision-language Transformers.
The main contributions of this work are summarized as follows:
\begin{itemize}
    \item We identify attention drift as a key robustness bottleneck in CIM-deployed ViTs and empirically establish its strong correlation with accuracy degradation under noise.

    \item We propose \textbf{ABNAT}, an attention-based KD framework that preserves semantic attention routes under CIM noise by aligning headwise attention maps between a clean teacher and a noisy student model.

    \item We introduce selective alignment variants at head and block granularity that reduce training cost while retaining accuracy close to full alignment.
    
    \item Extensive experiments spanning multiple noise levels, model scales, and datasets across vision, language, and multimodal tasks demonstrate that ABNAT consistently outperforms existing NAT baselines, achieving up to 11.28 percentage points higher noise-degraded accuracy and an 85.77\% reduction in attention deviation.
\end{itemize}

\section{Preliminaries}

\subsection{Transformers and Attention Mechanisms}

Transformers represent vision, language, and multimodal inputs as token sequences processed by $L$ stacked blocks~\cite{vaswani2017attention}. Each block combines multi-head self-attention (MHSA) and a feed-forward network with residual connections and layer normalization. Self-attention enables each token to aggregate relevant information from other permitted tokens, supporting content-dependent, long-range interactions. For head $h$ in block $l$, the input $\mathbf{X}_l\in\mathbb{R}^{N\times D}$ is projected into queries $\mathbf{Q}_{l,h}=\mathbf{X}_l\mathbf{W}^{Q}_{l,h}$, keys $\mathbf{K}_{l,h}=\mathbf{X}_l\mathbf{W}^{K}_{l,h}$, and values $\mathbf{V}_{l,h}=\mathbf{X}_l\mathbf{W}^{V}_{l,h}$. The attention map and head output are
\begin{equation}
\mathbf{A}_{l,h}=\operatorname{softmax}\!\left(\frac{\mathbf{Q}_{l,h}\mathbf{K}_{l,h}^{\top}}{\sqrt{d_k}}+\mathbf{M}\right),\quad \mathbf{O}_{l,h}=\mathbf{A}_{l,h}\mathbf{V}_{l,h}.
\end{equation}
Here, $d_k$ is the head dimension and $\mathbf{M}$ is an optional structural or causal mask. Each row of $\mathbf{A}_{l,h}$ defines how one query routes and aggregates information from the permitted tokens, while multiple heads capture complementary relations. Across blocks, these maps govern progressive information routing~\cite{abnar2020quantifying,clark2019attention}, making them a natural internal target for ABNAT. The same mechanism is shared by vision, language, and vision-language Transformers~\cite{dosovitskiy2021vit,biderman2023pythia,radford2021clip}.

\subsection{Analog CIM Hardware}

Analog CIM accelerators perform MVMs in situ by encoding weight matrices as analog device conductances and exploiting Ohm's law for multiplication and Kirchhoff's current law for summation~\cite{shafiee2016isaac, chi2016prime}. This in-memory computation eliminates costly data movement between memory and processing units, enabling orders-of-magnitude improvements in energy efficiency compared to conventional digital accelerators~\cite{sebastian2020memory, ielmini2018memory, chen2019cmosintegrated}. These properties make CIM an attractive substrate for energy-efficient Transformer deployment~\cite{li2023h3datten, misawa2025heterocim, li2025efficientedge}.

However, analog device non-idealities introduce unavoidable perturbations to the stored weights~\cite{burr2017neuromorphic, ambrogio2018equivalent}. These arise from multiple physical sources: conductance programming variation, thermal read noise, inter-cell interference, and long-term conductance drift. The detailed bit-level noise formulation used in our experiments is introduced in Sec.~\ref{sec:experiment}. For Transformers, noise corrupts the projection matrices $\mathbf{W}_Q$, $\mathbf{W}_K$, and $\mathbf{W}_V$ at every layer, perturbing attention computation throughout the network. Because Transformer representations are built compositionally across depth, these per-layer errors accumulate rather than cancel, causing severe attention drift in deeper layers where semantic concentration is most critical.

\subsection{Noise-Aware Training}

NAT improves CIM robustness by injecting hardware noise during training, optimizing:
\begin{equation}
    \min_{\mathbf{W}}\; \mathbb{E}_{\boldsymbol{\epsilon}}
    \bigl[\mathcal{L}_{\text{CE}}(f_{\hat{\mathbf{W}}}(\mathbf{x}), y)\bigr],
\end{equation}
where $\mathcal{L}_{\text{CE}}$ is the cross-entropy loss and $f_{\hat{\mathbf{W}}}$ denotes the noisy model. NAT has demonstrated effectiveness for CNNs~\cite{charan2020accurate, li2019vortex, zhou2020noisymachines}, but remains output-centric: supervising only final predictions places no explicit constraint on intermediate attention routes. The model may therefore learn shortcut solutions that produce correct, highly confident predictions through severely distorted attention maps. Such confidence--route inconsistency can generalize poorly across deployment noise levels and degrade clean accuracy because the semantic attention routes central to Transformer inference remain unconstrained.

KD transfers knowledge from a pretrained teacher $f_{\mathbf{W}}$ to a student $f_{\hat{\mathbf{W}}}$ by minimizing discrepancies in their outputs or intermediate representations, and has shown strong results in model compression and transfer learning~\cite{hinton2015distill,romero2015fitnets}. Prior studies have incorporated KD into NAT for CIM robustness: Charan et al.~\cite{charan2020accurate} distill the teacher's output predictions, whereas Zhou et al.~\cite{zhou2025rramvitkd} additionally align intermediate feature maps in RRAM-based ViTs. Despite using knowledge distillation, both approaches essentially teach the noisy student to reproduce answers while overlooking attention drift, without teaching it to understand the content under noise. In contrast, ABNAT transfers the clean teacher's attention distributions, teaching the noisy student to preserve content understanding under hardware noise. Section~\ref{sec:experiment} demonstrates the resulting gains in robust performance and cross-noise generalization.

\section{Methodology}


\subsection{Motivation}
\label{sec:motivation}

Let $f^T$ and $f^S$ denote the clean and noisy paths instantiated from the same pretrained ViT checkpoint. The clean path is frozen and serves as the teacher, while the noisy path injects CIM-induced weight noise and is optimized as the student; thus, ABNAT requires no separately trained teacher. Both paths contain $L$ encoder blocks and $H$ attention heads per block.
Given an input $x$, let 
$p^T_{l,h}(x), p^S_{l,h}(x) \in \Delta^{N-1}$ 
be the attention distributions over the $N$ patch tokens, excluding \texttt{[CLS]}, at block $l$ and head $h$ for the teacher and student, respectively. 
Each distribution is computed by averaging the post-softmax attention weights over all query tokens.

The \textbf{attention drift} of head $(l,h)$ is defined as
\begin{equation}
    \bar{d}_{l,h}
    =
    \mathbb{E}_{x}
    \left[
        D_{\mathrm{KL}}
        \left(
            p^T_{l,h}(x)
            \,\|\, 
            p^S_{l,h}(x)
        \right)
    \right].
    \label{eq:drift}
\end{equation}
A large $\bar{d}_{l,h}$ implies that CIM noise significantly changes the student's attention behavior from the clean reference, which may impair critical information flow. 
We therefore ask whether such drift is uniform across the network or follows structured patterns that can guide targeted training.

\begin{figure}[!t]
\centering
\includegraphics[width=\columnwidth]{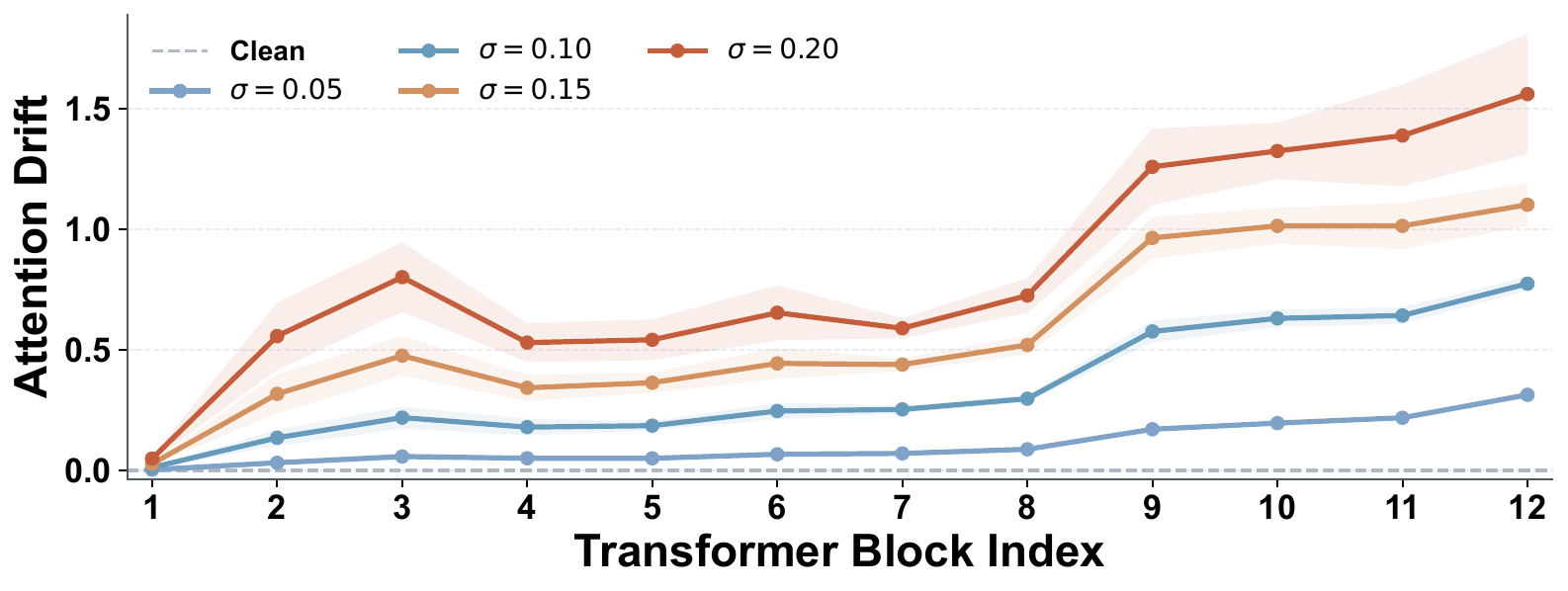}
\caption{Layerwise attention drift across CIM noise intensities; shaded bands denote std across seeds.}
\label{fig:layerwise_drift}
\end{figure}

As shown in Figure~\ref{fig:layerwise_drift}, $\bar{d}_{l,h}$ remains consistently low and nearly uniform across blocks~1--8, but undergoes a sharp, accelerating increase in blocks~9--12. This depth-dependent pattern can be understood through the lens of error propagation: perturbations introduced by CIM noise at any given layer are not confined to that layer, but are carried forward and progressively amplified as they pass through the subsequent nonlinear transformations of the Transformer. Consequently, late blocks, which receive as input the accumulated representations of all preceding layers, are disproportionately exposed to compounded noise effects. This observation has an important practical implication: a uniform attention alignment strategy that treats all blocks equally may be mismatched with the underlying structure of the problem. Instead, a \textbf{depth-aware} approach that concentrates alignment effort on late blocks is more efficient.

\begin{figure}[!t]
\centering
\includegraphics[width=\columnwidth]{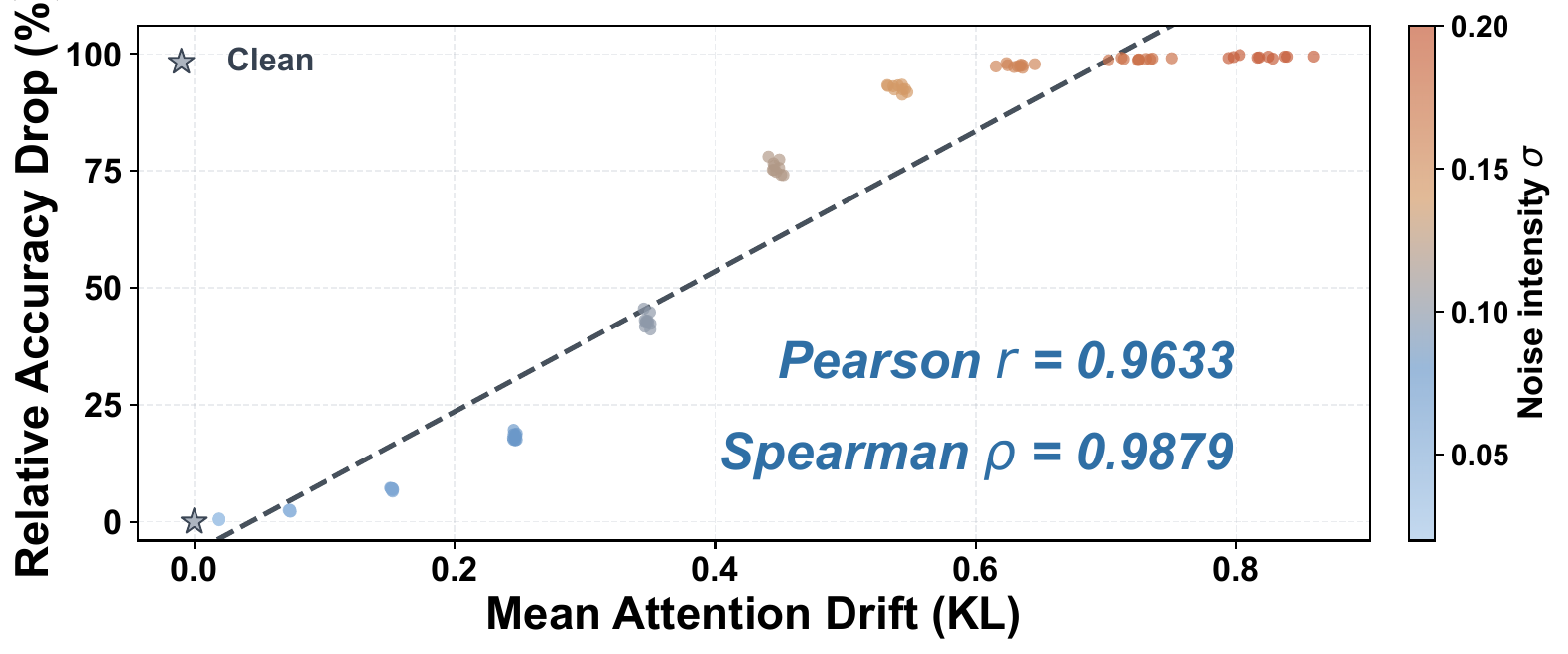}
\caption{Accuracy degradation versus mean attention drift across CIM noise settings.}
\label{fig:drift_correlation}
\end{figure}

The depth-dependent drift pattern indicates \emph{where} CIM noise has the strongest impact, but does not directly quantify its effect on task performance. 
To bridge this gap, Fig.~\ref{fig:drift_correlation} correlates the full-network mean attention drift with the model-level accuracy drop across diverse noise conditions and seeds. 
The results show a strong correlation, with Pearson $r = 0.9633$ and Spearman $\rho = 0.9879$.
This observation has two implications. 
First, mean attention drift serves as a reliable network-level robustness diagnostic: noise settings with larger mean drift are typically associated with greater accuracy degradation, whereas those with smaller drift exhibit less performance loss. 
Second, it identifies attention drift as a key robustness bottleneck under CIM noise, motivating our training framework to explicitly suppress drift rather than relying only on coarse output logits.

\subsection{ABNAT Framework}

\begin{figure*}[!t]
\centering
\includegraphics[width=0.95\textwidth]{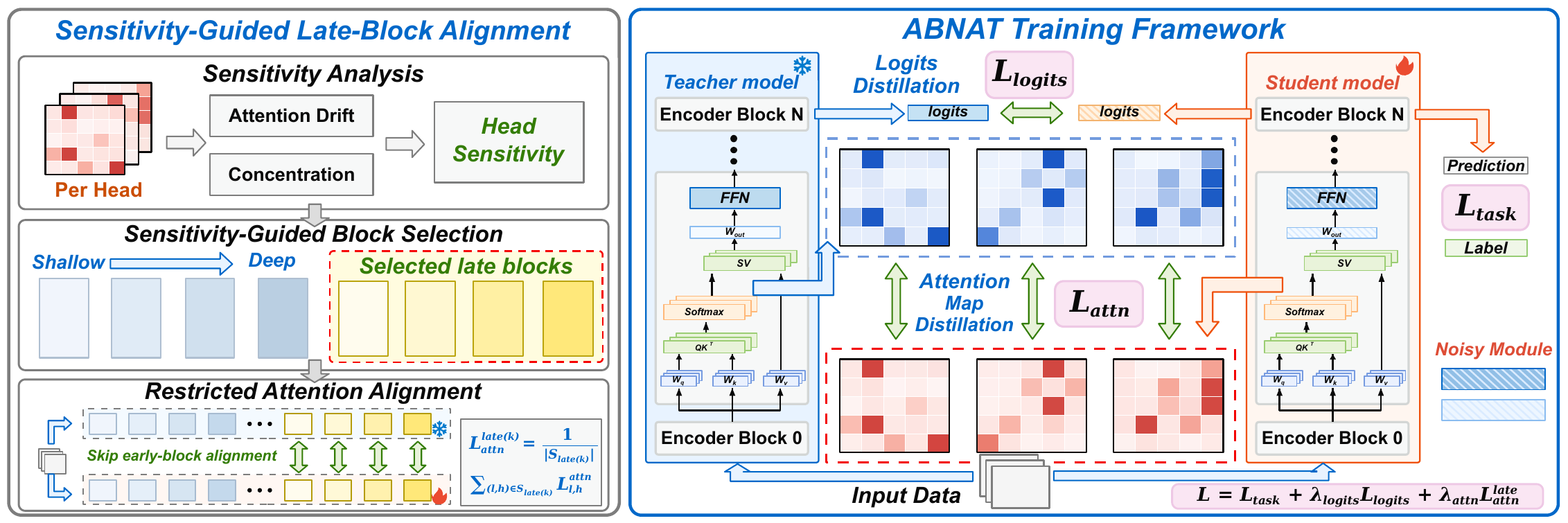}
\caption{ABNAT with sensitivity-guided late-block selection and task, logit-distillation, and attention-alignment objectives.}
\label{fig:framework}
\end{figure*}

The analyses above motivate full alignment for the strongest robustness and selective variants for lower training cost. Unless otherwise stated, ABNAT denotes full alignment over all attention heads and is used for the main accuracy results; \emph{ABNAT-selective} restricts the alignment set $\mathcal{S}$ defined below. Figure~\ref{fig:framework} summarizes the ABNAT pipeline. Although attention drift is concentrated in late blocks, both drift and clean attention concentration vary considerably across individual heads. To rank heads for sensitivity-guided selection, we therefore combine these two complementary factors. A concentrated head routes information through a small subset of tokens, whereas a diffuse head distributes attention broadly, making concentration a useful complement to drift when prioritizing routes for alignment.
\begin{figure}[!t]
\centering
\includegraphics[width=\columnwidth]{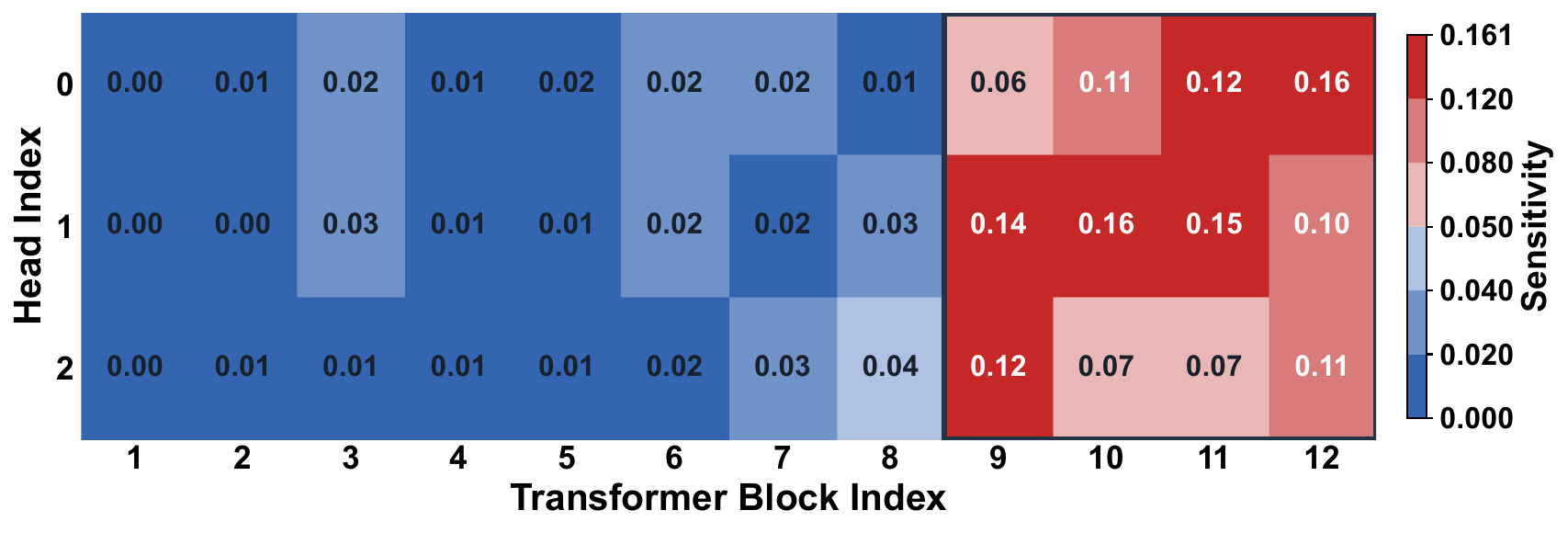}
\caption{Sensitivity scores $\mathrm{sens}_{l,h}$ across all Transformer blocks and attention heads.}
\label{fig:sensitivity_heatmap}
\end{figure}

We measure clean attention concentration as
\begin{equation}
    \bar{c}_{l,h} 
    = \mathbb{E}_x\!\left[
        1 - \frac{H(p^T_{l,h}(x))}{\log N}
    \right],
\end{equation}
where $H(p)=-\sum_{i=1}^{N}p_i\log p_i$ and $\log N$ is the maximum entropy. This normalization places $\bar{c}_{l,h}$ in $[0,1]$, with larger values indicating more concentrated attention. We compute it from the clean teacher so that it reflects the intended clean behavior independently of CIM noise.

The sensitivity score combines the two factors:
\begin{equation}
    \mathrm{sens}_{l,h} 
    = \bar{d}_{l,h} \cdot \bar{c}_{l,h}.
\end{equation}
A head therefore receives a high score only when it exhibits both large noise-induced drift and concentrated clean attention. A strongly drifted but diffuse head, or a concentrated but stable head, receives a lower score. The product thus ranks heads by jointly considering route disruption and clean attention focus, without treating either factor alone as sufficient. Figure~\ref{fig:sensitivity_heatmap} shows that high-sensitivity heads are concentrated primarily in blocks~9--12, with much lower scores in early blocks, reinforcing the depth-dependent pattern above.

\subsubsection{Sensitivity-Guided Head Selection}

We construct the \textbf{selected alignment set} $\mathcal{S}$ by retaining the $k$ heads with the highest sensitivity scores across all blocks and heads:
\begin{equation}
    \mathcal{S} 
    = \operatorname{Top-}k\!\left\{
        \mathrm{sens}_{l,h} \;\middle|\; 
        l \in \{1, \ldots, L\},\; h \in \{1, \ldots, H\}
    \right\}.
\end{equation}
Here, $k$ is a fixed training-overhead budget. For head-selective alignment, $\mathcal{S}$ is computed once before training by globally ranking all block--head pairs and is held fixed thereafter. Early-block heads remain eligible, while the observed concentration in late blocks emerges from the sensitivity ranking. Setting $\mathcal{S}$ to all heads recovers the default ABNAT.

\subsubsection{Attention Alignment over the Selected Set}

For a given $\mathcal{S}$, we define the attention alignment loss over its selected heads:
\begin{equation}
    L_{attn}(\mathcal{S})
    = \frac{1}{|\mathcal{S}|} 
      \sum_{(l,h) \in \mathcal{S}} 
      D_{\mathrm{KL}}\!\left(
          p^T_{l,h} \,\big\|\, p^S_{l,h}
      \right).
\end{equation}
The distributions used for alignment are obtained by dividing the corresponding attention logits by an attention temperature $\tau_{attn}$ before softmax; $\tau_{attn}=1$ recovers the standard attention distribution.
Minimizing this loss encourages the student to reproduce the teacher's clean attention patterns precisely at the heads identified as most critical, directly targeting the internal mechanisms through which CIM noise degrades performance.

No block is excluded a priori in sensitivity-guided head selection. In practice, high-sensitivity heads concentrate predominantly in late blocks, so $\mathcal{S}$ naturally focuses supervision on the most noise-vulnerable routes while retaining any high-scoring early heads. For the structured block-selective variants in Fig.~\ref{fig:lightweight_tradeoff}, $\mathcal{S}$ instead contains all heads within the selected blocks.

\subsubsection{Overall Training Objective}

Bringing all components together, the complete training objective of ABNAT combines three complementary losses, each operating at a distinct level of abstraction:
\begin{equation}
    L_{total}
    = L_{task}
    + \lambda_{logits}\, L_{logits}
    + \lambda_{attn}\, L_{attn}(\mathcal{S}),
\end{equation}
where $\lambda_{logits}$ and $\lambda_{attn}$ are scalar hyperparameters that balance the three objectives; $L_{task}$ is kept at unit coefficient as the scale anchor.
Unless otherwise stated, we use $\lambda_{logits}=3$, $\lambda_{attn}=50$, and $\tau_{\mathrm{KD}}=\tau_{attn}=1$.

At the coarsest level, the \textbf{task loss} $L_{task}$ is the standard cross-entropy loss computed against ground-truth labels, ensuring that the student retains discriminative performance on the target task regardless of noise conditions. Building on top of this, the \textbf{logits distillation loss} $L_{logits} = D_{\mathrm{KL}}(\mathrm{softmax}(z^T/\tau_{\mathrm{KD}}) \| \mathrm{softmax}(z^S/\tau_{\mathrm{KD}}))$, where $z^T$ and $z^S$ are the pre-softmax logits and $\tau_{\mathrm{KD}}$ is the distillation temperature, aligns the student's output probability distribution with that of the clean teacher. Finally, at the finest level of granularity, the \textbf{attention alignment loss} $L_{attn}(\mathcal{S})$ reaches deeper into the network to preserve the selected internal attention pathways. Together, these three losses form a coherent, multi-level training strategy: $L_{task}$ anchors the student to correct predictions, $L_{logits}$ aligns its output distribution with the teacher's, and $L_{attn}(\mathcal{S})$ safeguards attention routes vulnerable to CIM-induced degradation. This layered design allows ABNAT to simultaneously maintain high clean accuracy and achieve strong robustness under hardware noise.
\begin{table*}[!t]
\centering
\caption{Full component ablation for ViT-Tiny on CIFAR-100 under matched train--deploy noise. Accuracy under noisy inference is reported as mean$\pm$std over 30 seeds. Shaded rows denote the two KD-based ABNAT variants; ABNAT without $L_{task}$ retains logit KD and our attention-alignment loss.}
\label{tab:component_ablation}
\scriptsize
\setlength{\tabcolsep}{3.5pt}
\renewcommand{\arraystretch}{0.84}
\setlength{\aboverulesep}{0.30ex}
\setlength{\belowrulesep}{0.45ex}
\resizebox{\textwidth}{!}{%
\begin{tabular}{lccccccccc}
\toprule
Variant & Task loss & Logit KD & Feature~\cite{zhou2025rramvitkd} & Attn. Align. (Ours)
& $\sigma=0.05$ & $\sigma=0.10$ & $\sigma=0.15$ & $\sigma=0.20$ & Avg. \\
\midrule
Original+Noise
& -- & -- & -- & --
& 86.28$\pm$0.37 & 64.34$\pm$1.37 & 10.06$\pm$1.51 & 2.09$\pm$0.61 & 40.69 \\
NAT~\cite{rasch2023hardwareaware}
& $\checkmark$ & -- & -- & --
& 84.98$\pm$0.35 & 79.23$\pm$0.51 & 71.52$\pm$0.67 & 58.86$\pm$0.66 & 73.65 \\
KD
& -- & $\checkmark$ & -- & --
& 87.21$\pm$0.15 & 82.59$\pm$0.32 & 75.00$\pm$0.53 & 63.29$\pm$0.74 & 77.02 \\
FeatureAlign
& -- & -- & $\checkmark$ & --
& 86.46$\pm$0.28 & 80.36$\pm$0.50 & 43.27$\pm$1.33 & 14.72$\pm$1.59 & 56.20 \\
AttnAlign
& -- & -- & -- & $\checkmark$
& 86.22$\pm$0.29 & 78.68$\pm$0.49 & 54.76$\pm$1.14 & 12.91$\pm$1.21 & 58.14 \\
NAT+KD~\cite{charan2020accurate}
& $\checkmark$ & $\checkmark$ & -- & --
& 87.20$\pm$0.17 & 82.69$\pm$0.29 & 74.99$\pm$0.54 & 62.45$\pm$0.76 & 76.83 \\
NAT+Feature
& $\checkmark$ & -- & $\checkmark$ & --
& 86.49$\pm$0.20 & 81.67$\pm$0.43 & 71.87$\pm$0.95 & 58.35$\pm$0.72 & 74.60 \\
NAT+AttnAlign
& $\checkmark$ & -- & -- & $\checkmark$
& 87.15$\pm$0.21 & 83.37$\pm$0.41 & 77.01$\pm$0.70 & 66.05$\pm$0.92 & 78.39 \\
KD+Feature
& -- & $\checkmark$ & $\checkmark$ & --
& 87.64$\pm$0.20 & 83.22$\pm$0.36 & 75.53$\pm$0.62 & 63.15$\pm$0.80 & 77.39 \\
NAT+KD+Feature~\cite{zhou2025rramvitkd}
& $\checkmark$ & $\checkmark$ & $\checkmark$ & --
& 86.91$\pm$0.17 & 83.33$\pm$0.35 & 75.25$\pm$0.61 & 62.63$\pm$1.04 & 77.03 \\
\rowcolor{red!8}
ABNAT without $L_{task}$
& -- & $\checkmark$ & -- & $\checkmark$
& \cellcolor{green!12}\textbf{87.74$\pm$0.18}
& \cellcolor{green!12}\textbf{84.28$\pm$0.33}
& 78.62$\pm$0.56 & 70.04$\pm$0.82
& \cellcolor{green!12}\textbf{80.17} \\
\rowcolor{red!8}
\textbf{ABNAT}
& $\checkmark$ & $\checkmark$ & -- & $\checkmark$
& 87.62$\pm$0.21 & 83.95$\pm$0.28
& \cellcolor{green!12}\textbf{78.66$\pm$0.48}
& \cellcolor{green!12}\textbf{70.14$\pm$0.78}
& 80.09 \\
\bottomrule
\end{tabular}
}
\end{table*}

\begin{figure*}[!t]
\centering
\includegraphics[width=\textwidth]{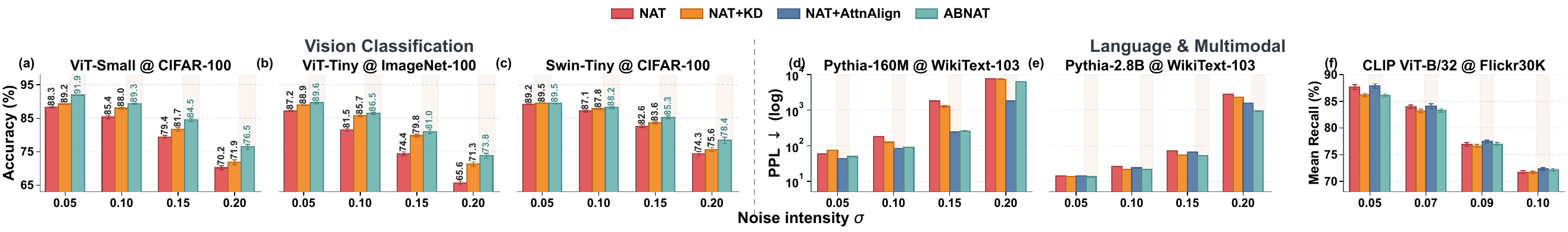}
\caption{Cross-setting results for ViT-Small on CIFAR-100, ViT-Tiny on ImageNet-100, Swin-Tiny on CIFAR-100, Pythia-160M and Pythia-2.8B on WikiText-103, and CLIP ViT-B/32 on Flickr30K.}
\label{fig:combined_generalization}
\end{figure*}

\section{Experiment}
\label{sec:experiment}
\subsection{Experimental Setup}
We evaluate ABNAT primarily on ViT-Tiny/16 with CIFAR-100, comparing NAT~\cite{rasch2023hardwareaware}, NAT+KD~\cite{charan2020accurate}, FeatureAlign~\cite{zhou2025rramvitkd}, and our AttnAlign.
Generalization is tested on ViT-Small on CIFAR-100, ViT-Tiny on ImageNet-100, Swin-Tiny on CIFAR-100, Pythia-160M and Pythia-2.8B on WikiText-103, and CLIP ViT-B/32 on Flickr30K.

Unless otherwise specified, models are trained with AdamW for 100 epochs, batch size 1024, learning rate $1\times10^{-3}$, weight decay 0.05, cosine decay, and 5 warmup epochs. 
For task-specific runs, the Pythia models use 262,144 WikiText-103 training tokens for 5 epochs at sequence length 128 and are evaluated on 65,536 tokens at length 2,048; CLIP uses Flickr30K training images at $224\times224$ for 10 epochs.
We use matched train--deploy noise, $\sigma_{\mathrm{train}}=\sigma_{\mathrm{infer}}$, and report mean $\pm$ std over 30 inference seeds. 
Unless otherwise stated, ABNAT denotes full alignment; Fig.~\ref{fig:lightweight_tradeoff} studies structured block-selective variants for lower training cost. 
Since alignment is training-only, ABNAT preserves the NAT inference cost.

CIM inference is modeled under W8A8 quantization with bit-level hardware variations, where $B_w=B_a=8$:
\begin{equation}
X_q = Q_a(X), \qquad W_q = Q_w(W),
\label{eq:quant}
\end{equation}
\begin{equation}
\begin{aligned}
X_q &= \sum_{j=0}^{B_a-1}2^jX^{(j)}, 
\qquad
W_q = \sum_{k=0}^{B_w-1}2^kW^{(k)}, \\
\widetilde{W}^{(k)}
&= W^{(k)} \odot 
\left(\mathbf{1}+\sigma\boldsymbol{\epsilon}^{(k)}\right),
\qquad
\boldsymbol{\epsilon}^{(k)} \sim \mathcal{N}(\mathbf{0},\mathbf{I}).
\end{aligned}
\label{eq:bitplane_noise}
\end{equation}
Here, $Q_a$ and $Q_w$ denote the activation and weight quantizers, $X^{(j)}$ and $W^{(k)}$ are the $j$-th activation and $k$-th weight bit planes, $\odot$ denotes element-wise multiplication, and $\sigma$ denotes the relative standard deviation of device-conductance variation.
For combined bit-parallel and bit-serial execution, the reconstructed CIM output is
\begin{equation}
Y_{\mathrm{CIM}}=
\sum_{j=0}^{B_a-1}\sum_{k=0}^{B_w-1}2^{j+k}
\Bigl[\Phi_{\mathrm{CIM}}\!\left(X^{(j)},\widetilde{W}^{(k)}\right)
+\boldsymbol{\epsilon}^{(j,k)}_{\mathrm{TIA\&ADC}}\Bigr],
\label{eq:cim_model}
\end{equation}
where $\Phi_{\mathrm{CIM}}$ is simulated by NeuroSim~\cite{read2025neurosim,peng2021dnn}, $\boldsymbol{\epsilon}_{\mathrm{TIA\&ADC}}^{(j,k)}$ denotes the aggregate readout error from the TIA and ADC, ADC precision is tied to the activated-row count, and mapped matrix operations use full-row activation.

\subsection{Experiment Results}

\noindent\textbf{Full Component Ablation}
Table~\ref{tab:component_ablation} ablates noisy task supervision, logit KD, feature alignment, and attention alignment under matched train--deploy noise. 
Our AttnAlign becomes increasingly effective as noise intensifies: adding it to NAT improves accuracy by 2.17 percentage points at $\sigma\!=\!0.05$ and by 7.19 points at $\sigma\!=\!0.20$. 
Together with logit KD, AttnAlign forms the complete ABNAT, which reaches 70.14\% at $\sigma\!=\!0.20$, outperforming NAT+KD and NAT by 7.69 and 11.28 points, respectively. 
In comparison, FeatureAlign provides negligible high-noise benefit: adding it to NAT changes accuracy from 58.86\% to 58.35\%, while adding it to NAT+KD changes 62.45\% to 62.63\%. 
ABNAT without $L_{\mathrm{task}}$ also achieves a strong average accuracy of 80.17\%, while complete ABNAT performs best at $\sigma\!\in\!\{0.15,0.20\}$. 
These results show that attention-route preservation becomes increasingly important as hardware noise intensifies.

\noindent\textbf{Generalization Across Models and Datasets}
We evaluate ABNAT's effectiveness across model architectures and datasets to assess generalizability. Figure~\ref{fig:combined_generalization}(a--c) presents results for ViT-Small on CIFAR-100, ViT-Tiny on ImageNet-100, and Swin-Tiny on CIFAR-100. At $\sigma\!=\!0.20$, ABNAT improves over NAT by 6.34, 8.15, and 4.15 percentage points, respectively. These consistent gains across model scales, datasets, and both flat and hierarchical Transformer architectures indicate that attention-route supervision is not tied to a single configuration.

\noindent\textbf{Generalization to Language and Multimodal Transformers}
We further evaluate attention-route preservation using Pythia-160M and Pythia-2.8B for language modeling~\cite{biderman2023pythia} and CLIP ViT-B/32 for image--text retrieval~\cite{radford2021clip}. Figure~\ref{fig:combined_generalization}(d--f) reports perplexity and Mean Recall, where lower and higher values are better, respectively. An attention-aligned variant performs best in all three settings. At $\sigma\!=\!0.20$, NAT+AttnAlign reduces Pythia-160M perplexity by 76.4\%, from 7682.16 to 1812.96, while NAT+KD+AttnAlign reduces Pythia-2.8B perplexity by 59.6\%, from 2329.72 to 941.04. On CLIP ViT-B/32 at $\sigma\!=\!0.10$, NAT+AttnAlign improves Mean Recall from 71.70 for NAT to 72.27. These results extend attention-route preservation to autoregressive language modeling and multimodal retrieval.

\begin{figure}[!t]
  \centering
  \includegraphics[width=\columnwidth]{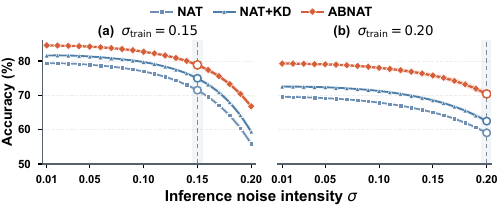}
  \caption{Train--deploy noise mismatch for ViT-Tiny on CIFAR-100 at $\sigma_{\text{train}}\!\in\!\{0.15,0.20\}$.}
  \label{fig:mismatch}
\end{figure}

\noindent\textbf{Robustness Under Train--Deploy Noise Mismatch}
Inference noise can deviate from training conditions because of device variation, lifetime drift, and operating changes~\cite{burr2017neuromorphic,ambrogio2018equivalent}. We therefore fix $\sigma_{\mathrm{train}}\in\{0.15,0.20\}$ and sweep $\sigma_{\mathrm{infer}}\in[0.01,0.20]$ in Fig.~\ref{fig:mismatch}. ABNAT outperforms NAT and NAT+KD throughout the sweep. For $\sigma_{\mathrm{train}}=0.15$, its gain over NAT reaches 10.86 points at $\sigma_{\mathrm{infer}}=0.20$; when trained at 0.20, it retains a 9.68-point gain even at $\sigma_{\mathrm{infer}}=0.01$. Under clean inference, the latter checkpoint maintains 79.29\% accuracy, compared with 69.59\% for NAT and 72.54\% for NAT+KD. These results indicate that attention-route supervision strengthens the noisy model's content understanding, enabling robust generalization across deployment noise while better preserving clean accuracy.

\begin{figure}[!t]
  \centering
  \includegraphics[width=\columnwidth]{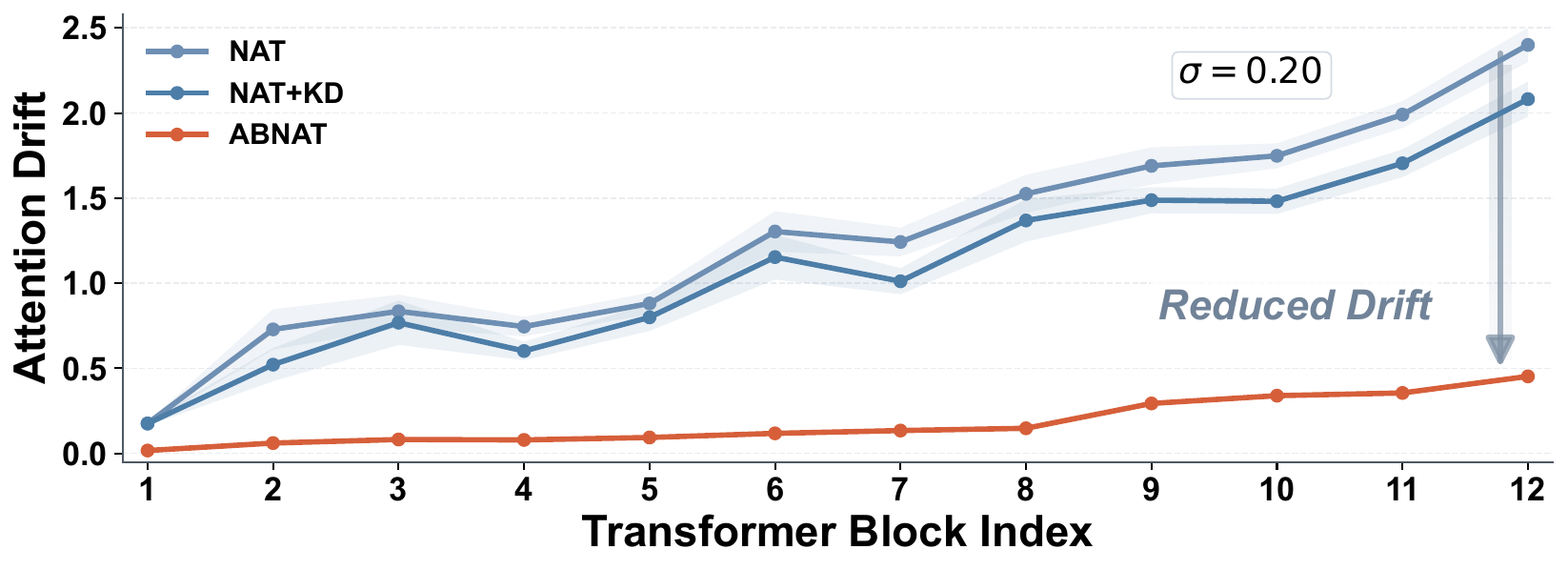}
  \caption{Layerwise attention drift of NAT, NAT+KD, and ABNAT at $\sigma\!=\!0.20$.}
  \label{fig:layerwise_drift_compare}
\end{figure}
\subsection{Discussion}
\noindent\textbf{Analysis of Attention-Route Stabilization}
We analyze attention drift patterns to understand ABNAT's performance improvements. Figure~\ref{fig:layerwise_drift_compare} shows that at $\sigma\!=\!0.20$, the block-averaged attention drift is 1.272 for NAT and 1.096 for NAT+KD, but only 0.181 for ABNAT---an 85.77\% reduction relative to NAT. ABNAT suppresses drift most strongly in the late blocks, where CIM noise causes the largest deviations. Together with the observed correlation between mean attention drift and accuracy degradation in Fig.~\ref{fig:drift_correlation}, these results support that ABNAT improves robustness by stabilizing clean attention routes rather than relying only on output-level supervision.
\begin{figure}[!t]
  \centering
  \includegraphics[width=\columnwidth]{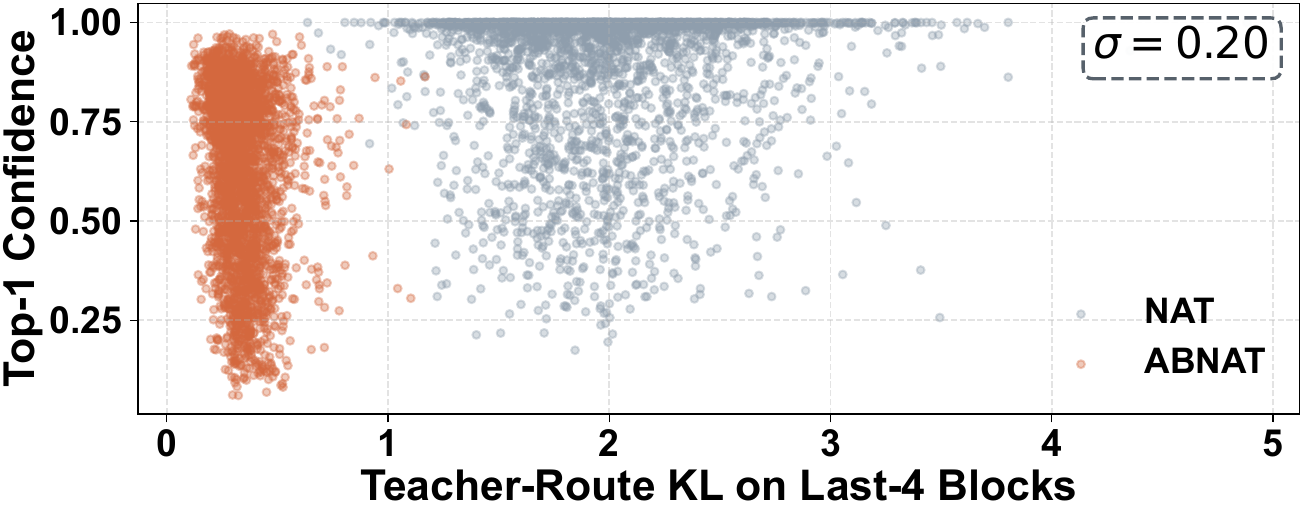}
  \caption{Prediction confidence versus attention-route fidelity at $\sigma\!=\!0.20$ on the common anchor subset.}
  \label{fig:confidence_route}
\end{figure}

\noindent\textbf{Prediction Confidence and Route Quality Analysis}
We analyze prediction confidence versus attention-route fidelity using the common anchor subset at $\sigma\!=\!0.20$, where NAT, ABNAT, and the clean teacher all predict correctly. Figure~\ref{fig:confidence_route} shows that NAT is more confident than ABNAT on 88.9\% of these samples, with a mean confidence of 0.899 versus 0.678. However, its teacher-route fidelity is substantially worse than ABNAT's, with a KL divergence of 1.972 versus 0.338 and a top-10 token overlap of 0.178 versus 0.360. This confidence--route inconsistency shows that output-focused training can produce highly confident predictions from noise-corrupted attention routes, whereas ABNAT more faithfully preserves the teacher's internal routing.

\begin{figure}[!t]
\centering
\includegraphics[width=\columnwidth]{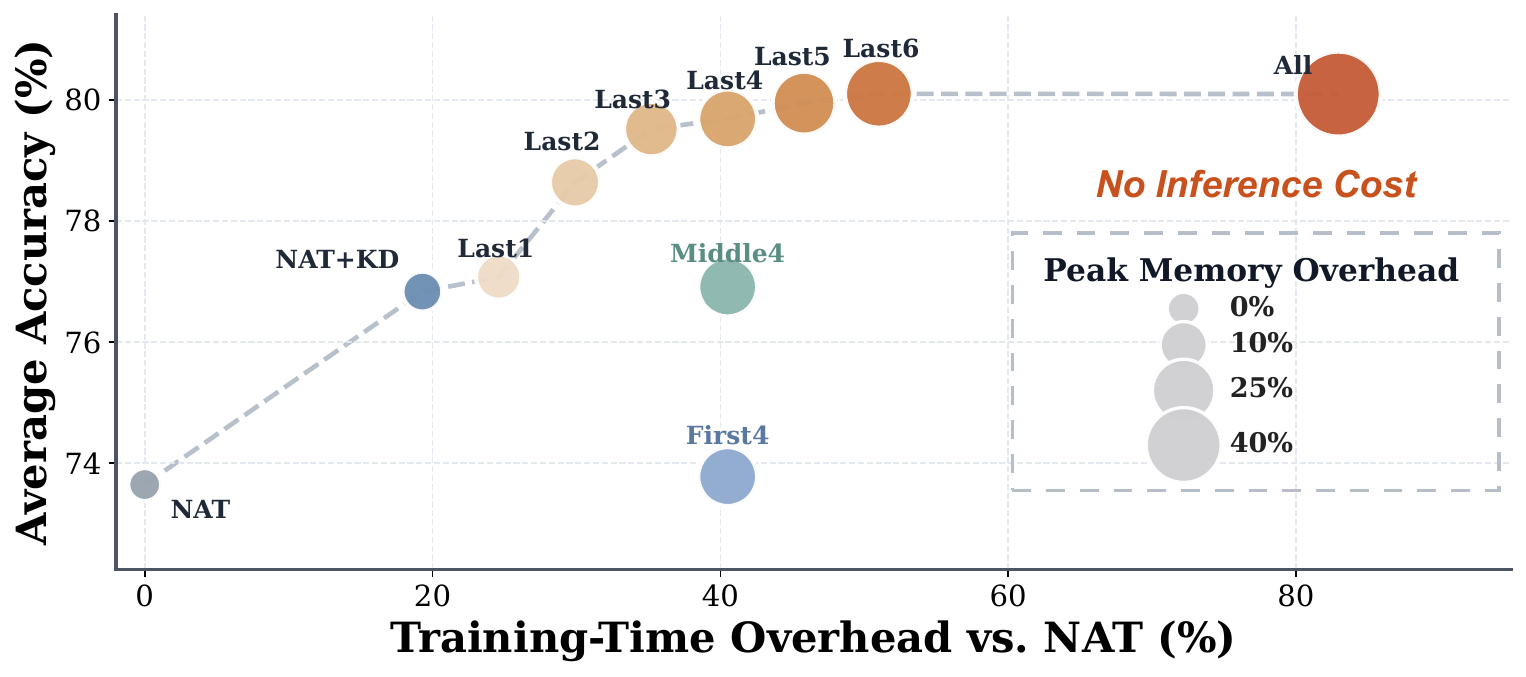}
\caption{Accuracy and training cost under selective block alignment.}
\label{fig:lightweight_tradeoff}
\end{figure}

\noindent\textbf{Accuracy--Training Cost Tradeoff}
Figure~\ref{fig:lightweight_tradeoff} evaluates structured block-selective variants against ABNAT with full alignment; here, $\mathcal{S}$ contains all heads in the selected blocks. Since attention alignment is training-only, all variants retain the same inference cost as NAT. Under the same four-block budget, Last4 outperforms First4 and Middle4, confirming that late blocks are more effective alignment targets. Accuracy improves and then saturates as the aligned late-block budget increases from Last1 to All. Last4 achieves 79.68\% average accuracy, only 0.41 percentage points below full alignment at 80.09\%, while reducing training-time overhead relative to NAT from 82.9\% to 40.8\% and peak-memory overhead from 54.4\% to 23.3\%. Thus, selective late-block alignment preserves nearly all robustness gains at substantially lower training cost.

\noindent\textbf{Head-Budget Sensitivity}
Head-selective alignment improves\linebreak smoothly with budget: average accuracy rises from 78.24\% to 78.90\% and 79.44\% as top-$k$ increases from 3 to 6 and 9 heads. At fixed $k=6$, high-drift heads achieve 78.90\% average accuracy, outperforming low-drift heads at 77.01\% by 1.89 percentage points. These results show that ABNAT is not sensitive to one particular budget and validate the sensitivity-guided selection.

\begin{table}[!t]
\centering
\caption{Hyperparameter sensitivity for ViT-Tiny on CIFAR-100. Bold marks the best setting in each sweep; shading marks defaults.}
\label{tab:hyperparam_sensitivity}
\scriptsize
\setlength{\tabcolsep}{1.0pt}
\renewcommand{\arraystretch}{0.84}
\setlength{\aboverulesep}{0.15ex}
\setlength{\belowrulesep}{0.25ex}
\begin{tabular*}{\columnwidth}{@{\extracolsep{\fill}}cccccccc@{}}
\toprule
\multicolumn{2}{c}{$\lambda_{logits}$} &
\multicolumn{2}{c}{$\lambda_{attn}$} &
\multicolumn{2}{c}{$\tau_{\mathrm{KD}}$} &
\multicolumn{2}{c}{$\tau_{attn}$} \\
\cmidrule(l{1pt}r{1pt}){1-2}\cmidrule(l{1pt}r{1pt}){3-4}\cmidrule(l{1pt}r{1pt}){5-6}\cmidrule(l{1pt}r{1pt}){7-8}
Value & Avg. & Value & Avg. & Value & Avg. & Value & Avg. \\
\midrule
0.25 & 78.41 & 1   & 78.52 & 0.5 & 79.36 & 0.125 & 67.57 \\
0.5  & 79.00 & 5   & 79.44 & \cellcolor{orange!10}1 & \cellcolor{orange!10}80.09 & 0.25 & 74.53 \\
1    & 79.52 & 10  & 79.85 & \textbf{2} & \textbf{80.17} & 0.5 & 78.47 \\
\cellcolor{orange!10}3 & \cellcolor{orange!10}80.09 & \textbf{25} & \textbf{80.16} & 3 & 79.90 & \cellcolor{orange!10}\textbf{1} & \cellcolor{orange!10}\textbf{80.09} \\
\textbf{5} & \textbf{80.19} & \cellcolor{orange!10}50 & \cellcolor{orange!10}80.09 & 4 & 79.80 & 2 & 79.93 \\
8    & 80.14 & 100 & 79.71 & 6 & 79.63 & 4 & 79.24 \\
12   & 80.05 & 200 & 78.69 & 8 & 79.46 & 8 & 78.37 \\
\bottomrule
\end{tabular*}
\end{table}

\noindent\textbf{Hyperparameter Sensitivity}
Table~\ref{tab:hyperparam_sensitivity} reports one-factor sweeps with all other settings fixed at their defaults. Around the defaults, $\lambda_{logits}\in[3,12]$, $\lambda_{attn}\in[25,100]$, and $\tau_{\mathrm{KD}}\in[1,4]$ retain 79.71--80.19\% average accuracy; $\tau_{attn}\in[0.5,4]$ yields 78.47--80.09\%, while the overly sharp value 0.125 drops to 67.57\%. Thus, ABNAT does not require delicate hyperparameter tuning.

\section{Conclusion}
\label{sec:conclusion}
In this paper, we show that CIM noise induces depth-accumulated, head-dependent attention drift, while output-centric NAT may reproduce answers through routes that deviate from the clean model. We propose ABNAT, which aligns headwise attention between clean and noisy paths from the same checkpoint to preserve task-relevant content processing rather than merely match answers. Across vision, language, and multimodal Transformers, ABNAT improves noisy accuracy by up to 11.28 percentage points, reduces attention drift by 85.77\%, generalizes under train--deploy noise mismatch, and adds no inference cost; selective variants retain near-full robustness with lower training overhead.

\begin{acks}
\setlength{\emergencystretch}{3em}
This work was supported by National R\&D Program of China (2022YFB4400300), NSFC (62534002, 62274003, 92564302) and 111~Project (B18001), in part by the STIC under Grant \mbox{QYJS-2022-1501-B} and the National Micro/Nano Fabrication Laboratory of Peking University.
\par
\end{acks}

\bibliographystyle{ACM-Reference-Format}
\bibliography{bibliography}
\end{document}